\documentclass[letterpaper, 10 pt, conference]{ieeeconf}

\usepackage{mathptmx}

\usepackage{appendix}

\usepackage{cite}
\usepackage{graphicx}
\usepackage{amsmath}
\usepackage{url}
\usepackage{algorithmic}
\usepackage{float}
\usepackage{wasysym}
\usepackage{threeparttable}
\usepackage{overpic}
\usepackage{psfrag}
\usepackage[dvipsnames]{xcolor}

\usepackage[normalem]{ulem}
\usepackage{tikz}

\IEEEoverridecommandlockouts

\title{The Robustness of Tether Friction in Non-idealized Terrains}

\author{Justin~J.~Page*,
        Laura~K.~Treers*,
        Steven~Jens~Jorgensen,
        Ronald~S.~Fearing,
        and~Hannah~S.~Stuart
    \thanks{J.J. Page, L.K. Treers and H.S. Stuart are with the Dept. of Mechanical Engineering, University of California at Berkeley, CA USA. S.J. Jorgensen is with the Johnson Space Center, NASA; Houston, TX USA. R.S. Fearing is with the Dept. of Electrical Engineering and Computer Sciences, University of California at Berkeley, CA USA. Correspondence to \tt{\footnotesize justin$\_$page@berkeley.edu}; \tt{\footnotesize hstuart@berkeley.edu}.}
    \thanks{* Shared first-authorship}
    \thanks{There is a video supplement associated with this work.}
    \thanks{This work has been submitted to the IEEE for possible publication. Copyright may be transferred without notice, after which this version may no longer be accessible.}
}

\begin{document}
\bstctlcite{IEEEexample:BSTcontrol}
\maketitle

\begin{abstract}
  Reduced traction limits the ability of mobile robotic systems to resist or apply large external loads, such as tugging a massive payload. One simple and versatile solution is to wrap a tether around naturally occurring objects to leverage the capstan effect and create exponentially-amplified holding forces. 
  Experiments show that an idealized capstan model explains force amplification experienced on common irregular outdoor objects -- trees, rocks, posts. Robust to variable environmental conditions, this exponential amplification method can harness single or multiple capstan objects, either in series or in parallel with a team of robots. This adaptability allows for a range of potential configurations especially useful for when objects cannot be fully encircled or gripped. 
  These principles are demonstrated with mobile platforms to \textit{(1)} control the lowering and arrest of a payload, \textit{(2)} to achieve planar control of a payload, and \textit{(3)} to act as an anchor point for a more massive platform to winch towards. We show the simple addition of a tether, wrapped around shallow stones in sand, amplifies holding force of a low-traction platform by up to 774x. 
  \\

\end{abstract}


\section{Introduction}

\begin{figure}[t]

    \centering
    \includegraphics[width=1\columnwidth]{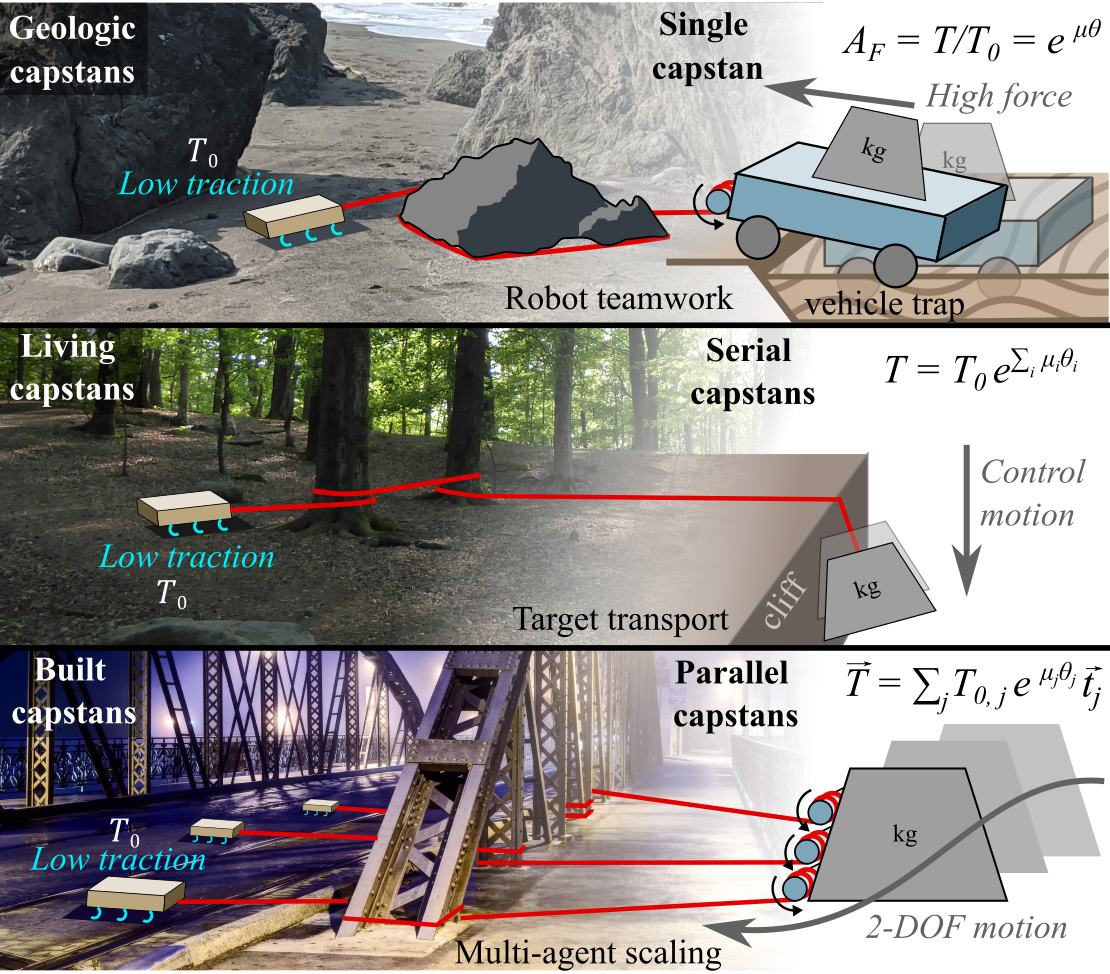}
    \setlength{\abovecaptionskip}{-10pt}
    \caption{ \textbf{Capstan-enabled maneuvers} Example exploitation of natural capstans for robotic missions on low-traction substrates are demonstrated. 
    Capstan objects are shown utilized in single (top), serial (middle), and parallel (bottom) configurations. Robotic teamwork, target transport, and multi-millibot or swarm scaling modes, respectively, are all enabled through capstan amplification. Photo credits: (top) H.S. Stuart, (middle) R. Henrik Nilsson, CC BY-SA 4.0, (bottom) Jar.ciurus, CC BY-SA 3.0 PL.
    }
    \label{fig:concept}
    \vspace{-6mm}
\end{figure}

Mobile robotic systems enable tasks like remote exploration, automated construction, and search and rescue. Natural terrains and unstructured environments can consist of loose and slippery media such as sand, shale, detritus, snow, mud, and biofilm. These surfaces support only limited shear and frictional contact forces for locomotion and payload maneuvering. Therefore, small robots often require enhanced or specialized foot designs to improve contact conditions, like 
treads, adhesives, or spines. Instead, we explore how to create forceful robot systems even when traction remains low. As depicted in Fig. \ref{fig:concept}, 
tethers connecting multiple robots are purposefully wrapped 
around terrain features available in the environment. 
 By taking advantage of fixed anchors, tether-world friction supports exponential amplification of ground traction forces and the effective load holding capacity of simple lightweight mobile robots.

\subsection{Background: Tethered robots for payload manipulation}
Prior work demonstrates how tethered teams of robots manipulate payloads \cite{estrada_forceful_2018,christensen_lets_2016} and function as exploratory agents tethered to a “mother-ship” \cite{mcgarey_system_2016, stefanini_tether_2018}. Systems overcome obstacles in irregular environments using robot-robot cooperation through pushing \cite{mcpherson_team-based_2019, deshpande_methodology_2009} and exploiting tethers \cite{fukushima_development_2001,asano_tethered_2010, kitai_proposal_2005}. Tethers enable the pulling of other robots or may act as static lines to provide paths for a locomotion network of spider-like robots \cite{shoval_design_1999, hao_design_2011}. Tethered demonstrations include pair-wise cooperation for stair climbing \cite{casarez_step_2016}, mapping extremely steep terrain \cite{mcgarey_developing_2018}, descending over steep edges \cite{bares_dante_1999,mumm_planetary_2004}, and larger scale ascent and descent with a tether \cite{schempf_self-rappelling_2009}. Single agent systems include a rover capable of tugging a sled on sand \cite{fernandez_tail-based_2021}, or a robot with “reaching arms” that moves a load when pulling in tension \cite{schneider2021reachbot}. 

These works assume robust attachment or traction with the world, achieved through specialized gripping or anchoring mechanisms \cite{estrada_forceful_2018,christensen_lets_2016,fernandez_tail-based_2021,schneider2021reachbot}, making an agent massive enough to be assumed unmovable \cite{mcgarey_system_2016}, or manually pre-deploying secure anchor points by researchers \cite{kitai_proposal_2005}. Applications outside of these specific scenarios warrant new robust and more generalized lightweight methods for the mobile creation of secure anchor points on the fly in the field. 

Many natural environments contain effectively fixed objects, such as trees or rocks. For tethered navigation in unstructured terrain, one work \cite{mcgarey_tslam_2017} provides tethered simultaneous localization and mapping (TSLAM) that estimates the location of tether-object contact points. Past literature \cite{stefanini_tether_2018,kumar_entanglement_2007,a_specian_friction_2015} cites these 
terrain features
as a primary disadvantage of tethered teams due to tether abrasion, unpredictable snagging, and tangling. In contrast, we leverage this tether-environment contact and propose to use these 
terrain
features to exponentially amplify the available resistive traction forces through the capstan effect. 

Despite the long-term establishment of capstans throughout history, the application to robotics is first mentioned in work by Augugliaro et al\cite{augugliaro_building_2013}, in which aerial vehicles are used to build tensile structures such as footbridges. This work focuses on the planning and execution of knots and wraps around indoor structures, expanded upon by Segal et al \cite{segal_method_2021}, who extend the method to natural features like trees. They do not characterize the variability or predictability of the simple capstan equation on natural objects like rocks and trees, nor do they discuss the amplification of contact forces as applied by mobile ground vehicles for flexible payload manipulation. 
Although complex models that account for tendon extensibility \cite{d_kuhm_fabric_2014} or radius and irregularity of capstan cylinder \cite{howell_friction_1954,jayawardana_capstan_2021} exist for controlled laboratory settings, there remains an opportunity to characterize the robustness of this mechanism on real-world terrains. 
\emph{We present the first study that verifies whether conventional formulations provide practical estimates for robot applications in the field, even with partial wraps on irregular natural capstan objects}.

\subsection{Overview}
Sec. \ref{sec:model} summarizes idealized tether friction amplification behavior. We contrast different robotic anchoring technologies with capstan amplification using scaling arguments for natural objects.
In Sec. \ref{sec:results}, we explore the applicability of the generalized capstan equation to multi-capstan systems in series, 
a previously unproven assumption. We then present 
experiments characterizing tether friction on natural objects in the field. 
We observe that capstan slip is often non-catastrophic and can further amplify holding force. 
In Sec. \ref{sec:demo}, we illustrate these principles on robotic systems in both the laboratory and the field, 
and Sec. \ref{sec:discuss} provides a discussion on real-world tether deployment considerations. Sec. \ref{sec:conclusion} summarizes applications for future work.

\section{Harnessing the Capstan Effect}
\label{sec:model}

Tether tension amplification factor, $A_F$, is the instantaneous ratio between load tension, $T$, and holding force, $T_0$, or that $A_F=T/T_0$. The capstan equation states that $A_F$ 
scales exponentially with both wrap angle, $\theta$, and the coefficient of friction between the tether and the capstan, $\mu$, or that $A_F=e^{\mu\theta}$. By harnessing the capstan effect, even lightweight machines on materials with low surface attachment strength, or low $T_0$, can withstand tremendous pulling forces $T$. 

As shown in Fig. 1, we envision small, lightweight, terrestrial mobile robots in single or multi-robot systems, capable of resisting large loads by applying capstan amplification on a range of geologic, living, or built features. A tethered team exploiting capstans can \textit{(i)} pull payloads, \textit{(ii)} lower payloads, and \textit{(iii)} achieve multi-dimensional motion. Additionally, the tethers can be \textit{(i)} wrapped around a single feature, \textit{(ii)} wrapped around multiple features, or \textit{(iii)} applied to multiple mobile robots in parallel. 
For systems with serial capstan wraps, we expect that, within the traditional capstan equation, the product of $\mu$ and $\theta$ may be summed within the exponential: $T = T_0e^{\sum_{i} \mu_i \theta_i }$. 
For systems with multiple mobile agents applying capstan wraps, individual tension vectors may be summed: $\vec{T} = \sum_{j} T_{0,j}e^{\mu_j \theta_j } \vec{t_j}$, where $\vec{t_j}$ represents the unit direction vector of each agent's tether.

\begin{figure}[t]

    \centering
    \includegraphics[width=1\linewidth]{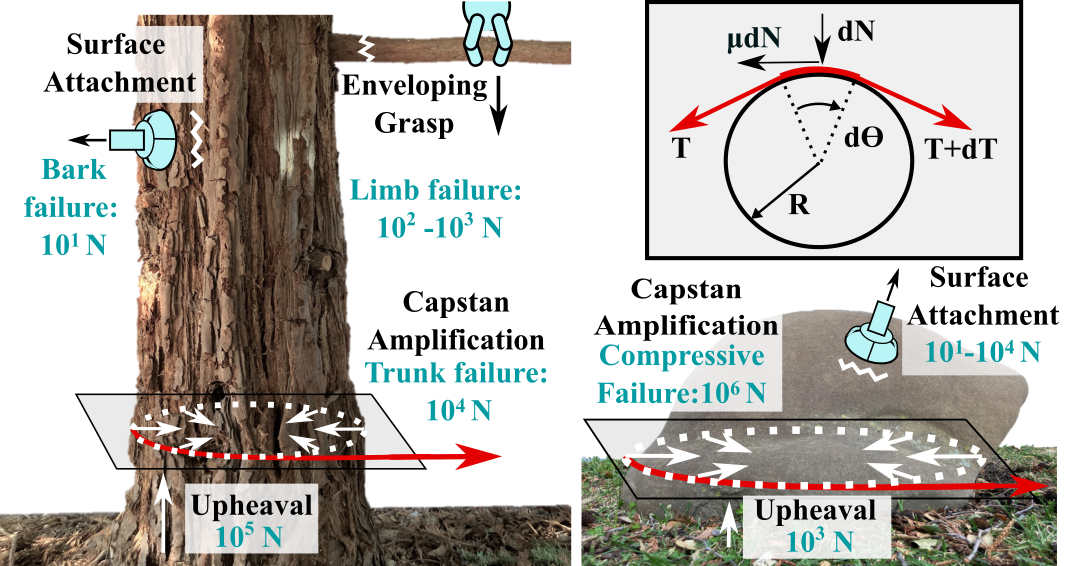}
    \setlength{\abovecaptionskip}{-10pt}
    \caption{\textbf{Comparison to robotic attachment methods.} Comparison of properties of various attachment modes frequently used by robots. 
    The simplified capstan effect amplifies tendon tension $T$ exponentially with wrap angle $\theta$ and friction coefficient $\mu$.  Tension scales linearly with normal force $N$ and is independent of local radius $R$.
    }
    \label{fig:comparison}
    \vspace{-6mm}
\end{figure}

\subsection{Comparison with other anchoring modes}

As shown in Fig. 2, we distinguish \textit{capstan amplification} from \textit{surface attachment} and \textit{enveloping grasps}. 
There is a rich history of \textit{surface attachment} mechanism design research, 
including 
spines \cite{jiang_stochastic_2018},  
suction \cite{yano_development_1997}, electrostatics \cite{prahlad_electroadhesive_2008}, and adhesives \cite{wang_large-payload_2013, hawkes_geckos_2013}. These astrictive grasping mechanisms develop high shear and normal forces at the surface, and thus fail when the surface material fails. They also tend to be specialized for specific surfaces; for example, present gecko-inspired adhesives perform exceptionally on clean polished surfaces, but reduce performance on rugose or dirty surfaces. \textit{Enveloping grasps} overcome these limitations by compressing the surface material, but require that the robotic gripper find features small enough to wrap around. 
Recent examples include a biomimetic robot for bird-like perching in arboreal environments \cite{roderick_bird-inspired_nodate}, autonomous perching by quadrotors on cylindrical objects \cite{thomas_visual_2016}, and a plant tendril-like robot that coils around branches and stiffens to create an envelopment grasp \cite{meder_plant_2022}.

In contrast, \textit{capstan amplification} with a tether allows for anchoring on objects of widely varying sizes and surface properties with greater strength. The capstan amplification effect is especially robust in its ability to exploit anchor objects with a variety of non-ideal surfaces, such as wet, crumbling, or rough surface conditions. Maximum object size is limited primarily by the length of tether available. At a sufficient wrap angle, object failure can occur under radial compression; in the case of a tree or rock, radial compression requires on the order of $10^4$ N or $10^6$ N of force, respectively \cite{green_wood_2001}\cite{wyllie_chapter_1996}, providing between two and five orders of magnitude more force than surface failure. 
On well-anchored  objects, i.e. force of upheaval is greater than maneuver force, tendon strength likely limits maximum loading ability before object failure occurs. Since the tether friction amplifies the tractive ability of an existing system, surface/enveloping methods can couple with capstans to create even stronger attachment forces by an agent. 

All attachment modes may fail if the feature itself moves: a tree is uprooted from the soil or a boulder slides along the ground. These upheaval forces vary with the object’s mass and ground interaction forces. For young trees and smaller rocks, upheaval forces range from tens to hundreds of Newtons, while for strongly rooted older trees and large boulders these forces reach the order of kN or larger \cite{lindroos_forces_2010}. 

\section{Experimentation of Variability}
\label{sec:results}


In order to characterize $A_F$ for the scenarios throughout this Section, we 
use a Mark-10 Series 4 hand-held force gauge. 
Data is obtained by 
first quantifying the holding force on a given substrate of a 25-gram 3D printed plastic sled, weighed down by a 200-gram calibration weight. A tether is tied to both the sled and the force gauge, then pulled via the force gauge until the sled’s first instance of slip, at which peak force is recorded as $T_0$. This process is repeated to identify the tension required to induce tether slip at a certain wrap angle, $\theta$, by 
first winding the tether around the capstan(s). 
Throughout each of these trials, we keep tether length, angle, and contact height constant. 
Unless specified, the tether used in all experiments is HERCULES Dyneema 1mm, a pre-braided and abrasion-resistant fishing line. 

\subsection{Scaling to multiple capstans}
An extension of the simple capstan equation to multi-capstan systems predicts equal forces for equal total wrap angles, regardless of the number of capstans involved, provided the friction coefficient is constant on all capstan objects. 
As shown in Fig. 3, a laboratory test bed consists of four one-inch (25mm) diameter steel rods on a Radiata pine plywood platform. Each rod is wrapped in 180-grit sandpaper 
to mimic a tether-rock 
interface both in friction and abrasive qualities. The tether and sandpaper 
were replaced frequently to diminish changes in the coefficients of friction between worn surfaces. With this setup, the number of capstans, path of the tether, and wrap angles can all be varied.
The tether friction of each cylinder is individually measured with a 720-degree wrap. 
The variation between the highest and lowest $A_F$ is 32$\%$, with mean friction coefficient 0.6$\pm$0.1.

As shown in Fig. 3A1(a)-(j), we measure holding force amplification across different \textit{permutations of serial wraps}, 
representing total summed wrap angles of 360 or 720 
degrees across one, two, or four cylindrical rods. The resulting $A_F$ at first instance of slip is measured for 15 trials at each combination of conditions. 
In Fig. 3C, which represents data from 3A combined across all 720 and 360- degree trials, $A_F$ at slip is significantly higher for a single capstan wrap than in cases with multiple capstans in series, at 36$\%$ and 29$\%$ higher, respectively. 
Statistical significance is tested with 
a one-way ANOVA test 
followed by Tukey’s HSD.

\begin{figure}[t!]
    \centering
    \includegraphics[width=1\linewidth]{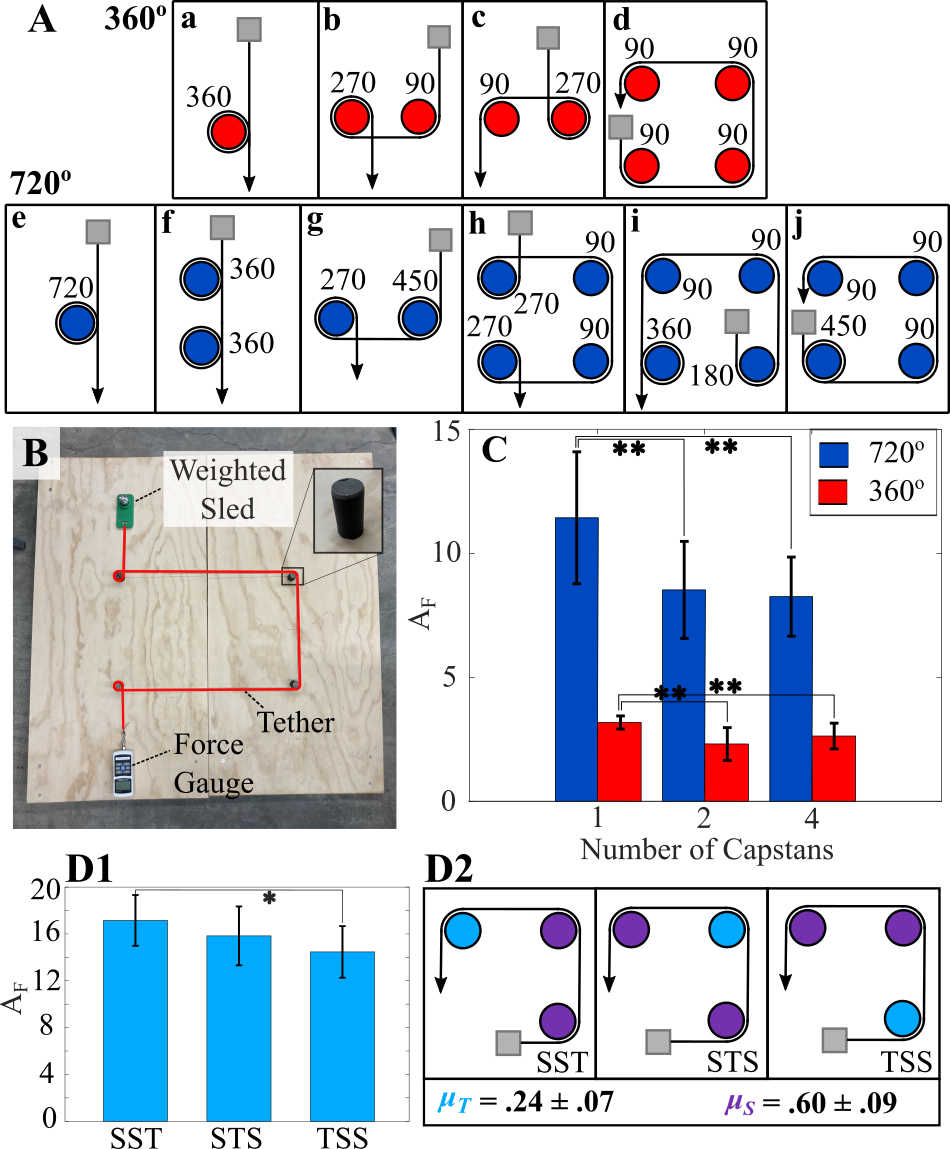}
    \setlength{\abovecaptionskip}{-10pt}
    \caption{ \textbf{Investigation of capstan effect across multiple capstans.} \textbf{(A1a-j)} All tested permutations of wrap angles summing to 360, 720, and 900 degrees, corresponding to data in (C). 
    \textbf{(B)} Experimental setup for laboratory capstan testing
    \textbf{(C)} Mean $A_F$ at slip produced as a function of number of capstans. 
    Statistical significance is indicated with ** corresponding to p $<$ 0.005.  
    \textbf{(D1)} $A_F$ for the various permutations of capstan material order presented in \textbf{(D2)}. S and T abbreviate sandpaper and tape, their friction coefficients $\mu_S$ and $\mu_T$. * corresponds to p $<$ 0.05.
    }
    \vspace{-6mm}
    \label{fig:no. capstans}
\end{figure}

The discrepancy in $A_F$ observed between one and multi-capstan wraps can be explained by the sensitivity of the capstan equation to small changes in friction coefficient, $\mu$, at large wrap angles. According to the capstan equation, differentiating $A_F$ with respect to both $\mu$ and $\theta$ yields $\frac{d}{d\mu} A_F = \theta e^{\mu\theta}$ and $\frac{d}{d\theta} A_F = \mu e^{\mu\theta}$. The typical $\mu$ is on the order of 0.25 - 0.5 for natural objects and 0.24 for the tape used on the lab testbed. Typical wrap angles for practical applications will be significantly larger than 0.5 radians or $\sim$30 degrees. Thus, we note that $\theta > \mu$ and ergo $\frac{d}{d\mu} A_F > \frac{d}{d\theta} A_F$. 
For a single capstan wrap, and assuming a typical $\mu$ of 0.3 and $A_F$ of 10, a 1\% change in $\theta$ will result in a 3.7\% change in $A_F$. However, for an $A_F$ of 10 with a wrap angle of 720 degrees, a 1\% change in $\mu$ will result in a 125\% change in $A_F$. 
As wrap angles increase, $A_F$ becomes more sensitive to small changes in $\mu$. 

We test the role of variable friction (Fig. 3D) by placing a lower friction capstan material (3M gaffers tape), 
among a 720-degree wrap containing two other capstans of a higher friction material (180-grit sandpaper).
The sequence which places the lowest friction capstan closest to the holding force (Tape Sandpaper Sandpaper or TSS) demonstrates the lowest mean force amplification. Likewise, placing the lowest friction capstan closest to the load (SST) results in the highest mean force amplification. We observe statistical significance between these two extreme scenarios, but not when they are compared with the intermediate scenario (STS). Given our friction sensitivity analysis and large variation observed, $A_F$ at slip is assumed to be independent of the sequence of $\mu$. 

As expected, this data supports the idea that the product of $\mu$ and $\theta$ may be summed within the exponential for multi-capstan wraps with varying friction. 
In practice, multiple capstans are preferred over a single capstan; 
when objects present a variety of friction coefficients that are difficult to predict accurately, 
multiple capstans provide an averaging effect that is insensitive to object order. However, if the highest friction object is identifiable, then a single capstan utilizing that one object supports higher tether tensions for a given wrap angle.

\subsection{Capstans in the natural environment}

We measure the amplification factor of different outdoor objects – geologic, living, and built – in order to provide the first investigation of whether the simple capstan effect holds true across a variety of real objects in the field. We select London planetrees (n=2), a palm tree, rocks (n=4), a lamp post, a fire hydrant, and redwood trees (n=10) at the University of California at Berkeley campus to provide a variety of common local objects. 
For each object, we wrap the tether to 90, 180, 270, 360 or 450 degrees and record the resulting tension required to induce tether slip; each condition is repeated five times. Snags on the bark are avoided during tether application. 
$A_F$ at slip is computed across these trials using a $T_0$ value collected over five independent trials at the location of each object, without any object wrapping or tether friction. 

\begin{figure}[t!]
    \includegraphics[width=1\linewidth]{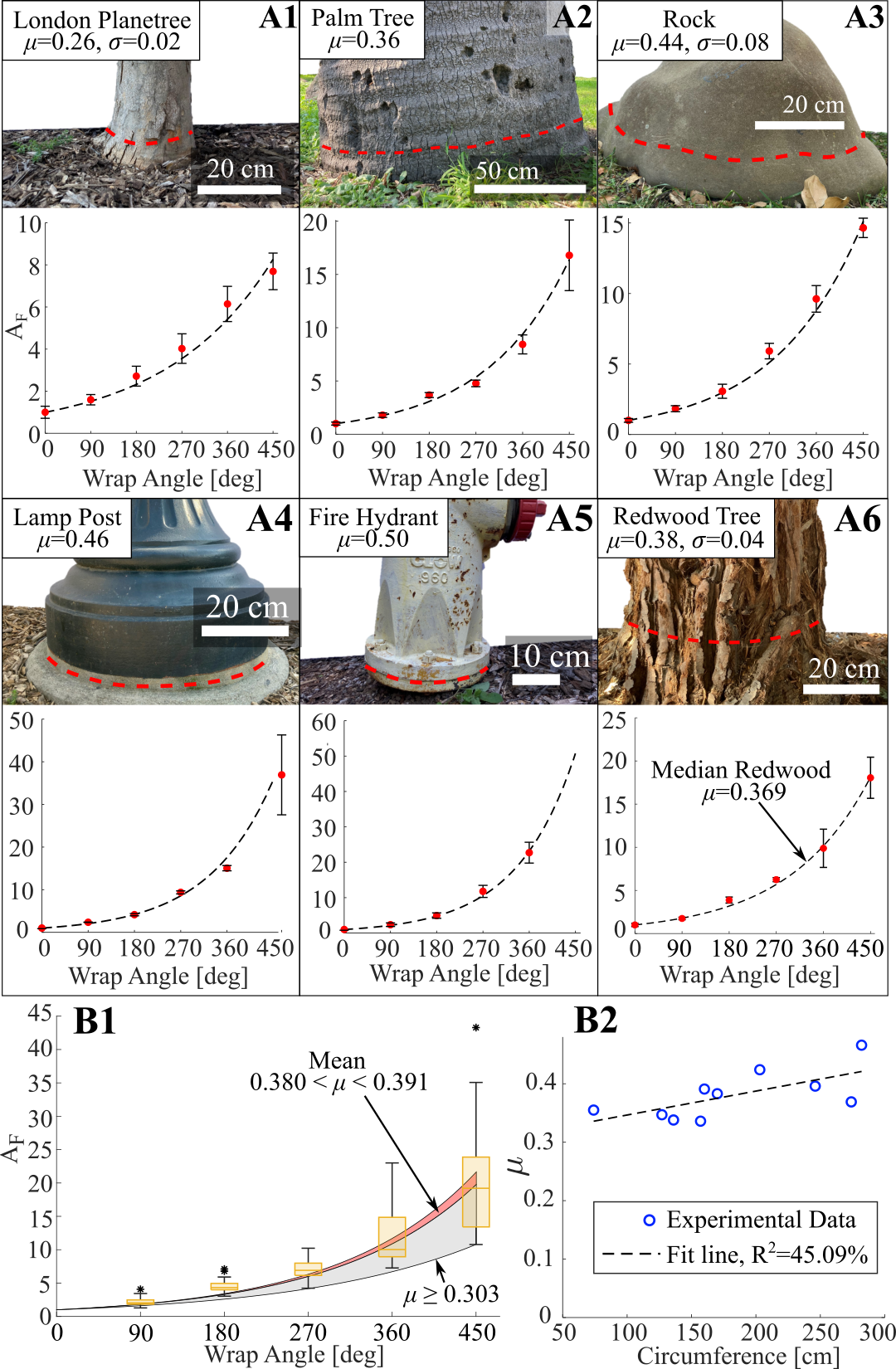}
    \setlength{\abovecaptionskip}{-10pt}
    \caption{ \textbf{Characterization of friction for various natural capstans.} \textbf{(A1) - (A6)} Images of capstan objects, with friction coefficient fit to the means of 5 trials at each wrap angle. Wrapped tether location indicated with a dashed red line. Standard deviation is reported for multi-object data sets. Plots below each image show the measured mean and standard deviation tension amplification at each wrap angle with a dashed curve representing the fit $\mu$. \textbf{(B1)} Tension amplification measured across 10 redwoods and 5 different wrap angles, corresponding to 50 data points at each wrap angle, described by the box plots. The red shaded region indicates the 95\% confidence interval on the friction coefficient fit to all data points. The lowest black line and gray region above indicates the lowest measured friction coefficient for wrap angles above 360 degrees.  \textbf{(B2)} Plot of fit friction coefficient for each redwood as a function of tree circumference, with a fit line that indicates a weak positive correlation. }
    \label{fig:muinnature}
    \vspace{-6mm}
\end{figure}

As reported in Fig. 4A1-6, the coefficient of friction µ in the capstan equation is treated as a fit parameter, and the exponential curve is fit to five data points using the MATLAB R2019b Curve Fitting toolbox. 
It should be noted that coefficients of friction may depend on the ever-changing wear, aging, abrasion, and interactions of particles sheared off of both the tether and capstan object over many trials; we reduce environmental variability by conducting experiments on days with similar conditions and replacing the tether between trials. Further, all data for each individual capstan object was gathered in one day. The coefficients of friction of the objects shown in Fig. 4A1-6 range from 0.26 for smooth bark to 0.5 for the fire hydrant. The $A_F$ versus wrap angle relationship of each of these objects follows an exponential trend matching the capstan effect, despite deviation from the idealized smooth and circular geometry, e.g., the redwood tree is highly rugose, and the rock is irregularly shaped.

We investigate intraspecies variation by comparing the 10 different \textit{Sequoia sempervirens} (coast redwoods) in order to inform expected performance of capstan amplification in specific environments; the current data set is particularly useful in a redwood forest. Coefficients of friction range from 0.336 to 0.466 with mean $\mu$= 0.38 and $\sigma$ = 0.04, as reported in Fig. 4B1. 
Shaded regions represent the fit and confidence bounds over all trees. 
We propose that an agent can estimate amplification factors, with an associated level of confidence, through visual classification of available object types alone, which is useful for planning maneuvers.

The idealized capstan equation neglects the shape and diameter of the capstan and amount of surface contact between the tether and capstan. Prior literature shows little effect of object radius on corresponding capstan force in the lab \cite{howell_friction_1954}. Circumferences of natural capstans of the \textit{Sequoia sempervirens} are estimated using strands of rope wrapped in an identical fashion to that in force experiments, and later measured against a tape measure. We check for diameter correlation across the \textit{Sequoia sempervirens} data set in Fig. 4B4, where we plot the mean calculated friction coefficients for each redwood tree as a function of its circumference. Only a weak positive correlation is present, largely supporting the diametral independence assumption for this species. The slight trend may be rooted in the rugosity of this species’ bark. Larger-radius natural objects provide a larger surface area, and thus increase the chance of imperfections and snag points on the tether that increase $A_F$. Younger redwood trees tend to have smoother bark and smaller trunk diameters. 

In the present work, we assume dry conditions. However, weather conditions may affect the friction coefficients of tether-object contact. We collect initial data on the experimental test bed (Fig. 3B); two different strings are wrapped in both dry and wet conditions around an object at 90-degree increments up to 450 degrees. The objects selected are a steel metal post, a post covered in tape as in Fig. 3, and a rock. We introduce dampness into the tether-object system via soaking the object and tether. Wetness did not consistently increase or decrease the coefficient of friction. For HERCULES Dyneema 1mm, the change in friction coefficient from dry to damp was +13.0\% for the tape post, -0.9\% for the metal post, and -1.5\% for the rock. For PTFE yarn, the changes were +5.6\%, +7.3\%, and +10.0\%.  
Thus, environmental-capstan appears robust to moisture. 


\subsection{Exceeding the capstan effect: Non-catastrophic slip}

Once a system’s applied loading force exceeds the capstan-amplified holding force, slipping of the tether on the capstan occurs. In real-world applications, forces can increase after this onset of tether slip such that these capstan failures are not catastrophic for the maneuver. We observe that slip can \textit{(i)} induce geometric snagging of the tether on the capstan and \textit{(ii)} increase the $T_0$ by pulling the agent through the terrain.

As greater force is applied to a tether, it cinches, leading the tether to settle in local concavities on the capstan object. As slip occurs, the tether can further dig into capstan objects and enter a pinched position. As shown in Fig. 5A and 5B, the tether becomes pinned beneath bark features on a tree or within crevasses of a jagged rock. Snags can appear immediately upon loading or develop after tether slipping, yet Fig. 5A and 5B both show the case in which snagging occurs after initial slip. Snagging increases the amplification factor by approximately 3x in both cases shown. We hypothesize that the shear strength of the capstan-object’s surface correlates with snagging amplification factor, as a weak surface would shear off under snagging stresses. While an advantageous mechanism for increasing holding force, snagging is disadvantageous if the tethered agent is intended for repeatable use, as snagging can impede unwinding.

As an object is dragged across a loose substrate, particles gather on the leading surface over time, and this interaction increases the $T_0$ capabilities of the agent. Heterogeneous substrates, such as the mulch in Fig. 5A2, contain sticks and other particles that can build up resistance to an object dragged through the surface. As a result, stick-slip force variations of the holding agent increase; as seen in Fig. 5C2, the sliding force on mulch varies by 70$\%$ as compared with 38$\%$ on smooth plywood. In a more homogeneous media, like sand, $T_0$ increases until granular resistive forces reach a steady state \cite{li_terradynamics_2013}. One pull test in a bed of granular media shows an increase of holding force by 33$\%$ due to sand mounding, as seen Fig. 5C3.

Each of the above phenomena compound to assist in arresting tether slip events. 
Robots may therefore operate close to capstan limits with low risk.
Future system designs may amplify these holding effects, for example by actively increasing mounding. 
It must be noted that slip wears tethers, thus this modality can fail over time due to tether abrasion.


\begin{figure}[t!]
    \centering
    \includegraphics[width=1\linewidth]{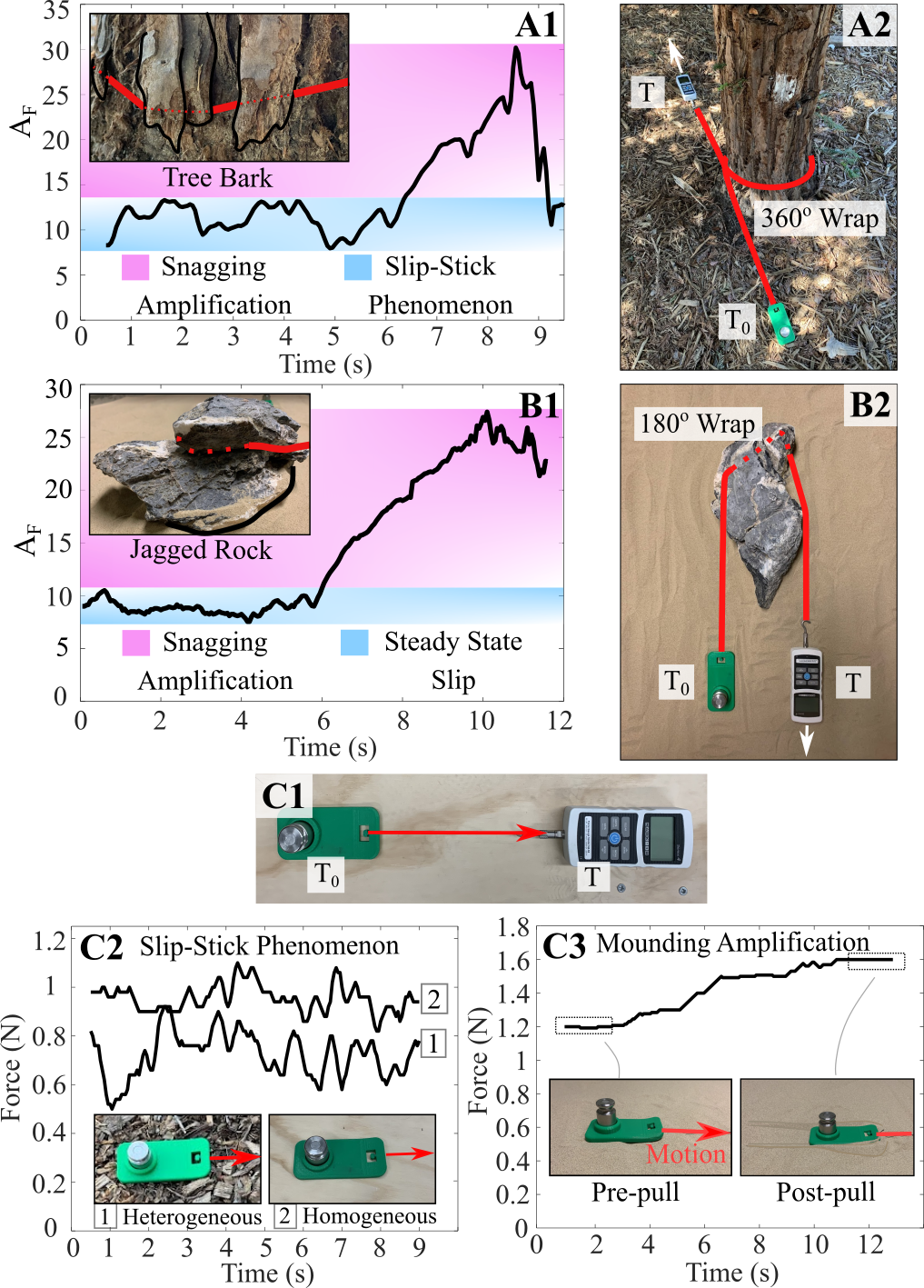}
    \setlength{\abovecaptionskip}{-10pt}
    \caption{\textbf{Dynamic slipping of capstan-tether system.} 
    \textbf{(A1)} A tether wrapped around a redwood tree is manually pulled until slip and continued to be pulled for approximately 10 seconds. $A_F$ over time is plotted. 
    \textbf{(A2)} Experimental setup for redwood tree snagging, including a weighted sled ($T_0$), handheld force gauge ($T$), and a 360 degree wrap of the tree. \textbf{(B1)} A tether wrapped around a rock is pulled for 12 seconds after initial slip and $A_F$ over time is plotted. \textbf{(B2)}  Experimental setup for rock snagging data. \textbf{(C1)} Experimental setup for recording slip-stick and mounding phenomena, in which holding force is measured directly without capstan amplification. \textbf{(C2)} A sled is dragged across a heterogeneous and homogeneous substrate, experiencing variation in force required to promote motion. \textbf{(C3)} A sled dragged through one meter of MARS90 regolith simulant shows a gradual increase and subsequent leveling of force required to slip over time.
     }
    \label{fig:failuremodes}
    \vspace{-6mm}
\end{figure}

\section{Robotic Demonstration}
\label{sec:demo}

For robotic demonstrations, we use a MiniRHex as a representative small-scale tether-deploying agent. This cockroach-inspired robot employs six semi-compliant, 360-degree-rotating semicircular ``C-shaped'' legs shown to overcome loose natural terrain \cite{barragan_minirhex_2018} (weighs .425 kg and measures .19m long by .10m wide by .10m tall). 
A separate payload winching platform uses a 
75:1 geared brushed DC motor coupled to a spool for reeling and retaining the tether, affixed to an acrylic sheet. This platform is then attached to three separate bases: caster wheels for hard surfaces, a sled for granular surface, or a rover. 
The caster and sled dollys weigh 3.34 kg  and 2.62 kg, respectively.
The rover, from \cite{cao_mobility_2021}, is 4.8 kg, 0.30m long by 0.30m wide by 0.15m tall. 


\begin{figure}[t!]
    \centering
    \includegraphics[width=1\linewidth]{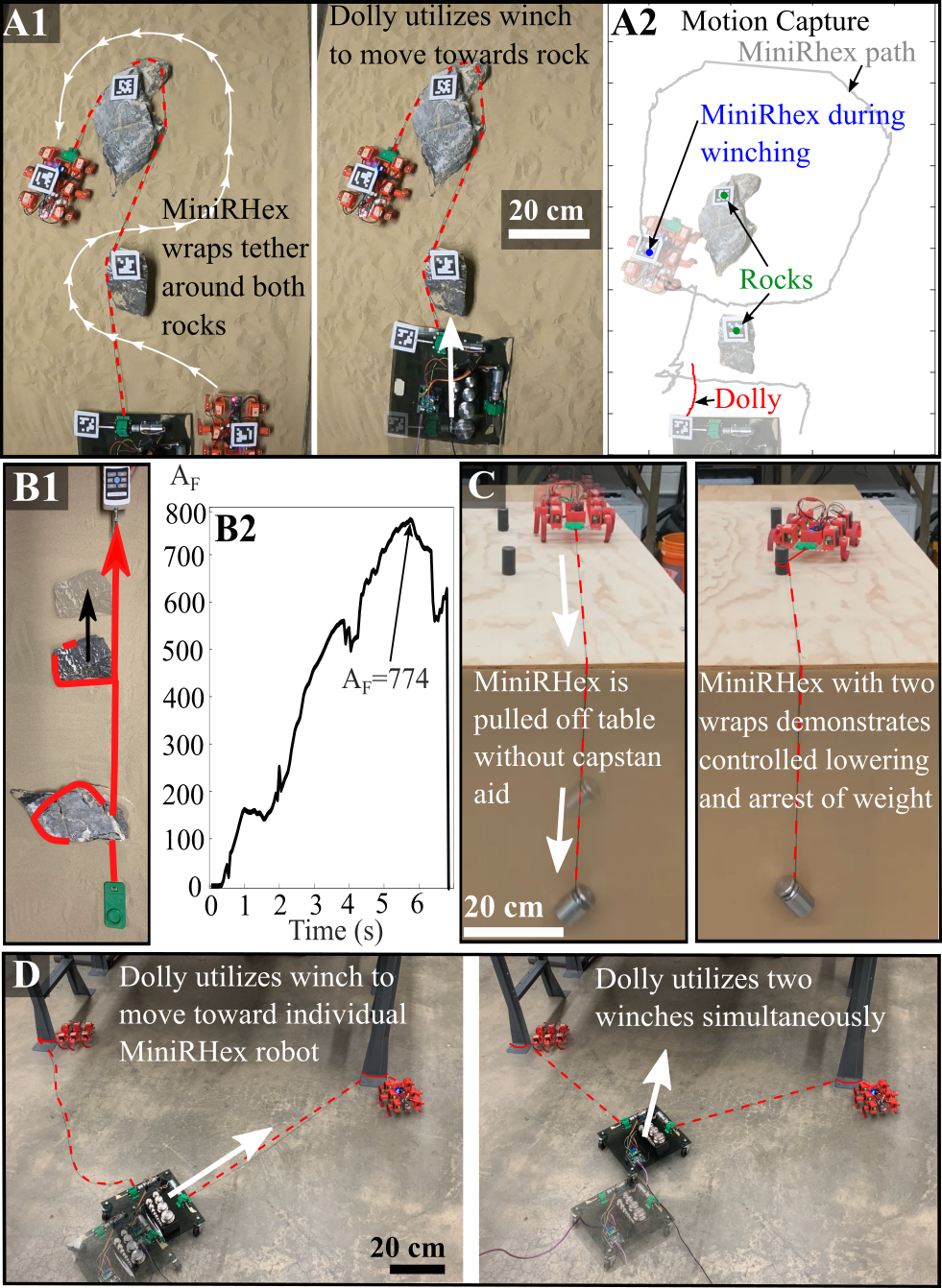}
    \setlength{\abovecaptionskip}{-10pt}
    \caption{ \textbf{Visual summary of laboratory robotic demonstrations} \textbf{(A1)} MiniRHex in granular media applies partial tether wraps on two rock capstan objects. A dolly equipped with a winch then winches towards the anchored MiniRHex. \textbf{(A2)} Motion capture data over time of the rocks, dolly, and MiniRHex. \textbf{(B1)} The same two rock capstans as in (A1) are wrapped to 360 degrees and pulled until failure. 
    \textbf{(B2)} $A_F$ over time is plotted, demonstrating a peak $A_F$ of 774. \textbf{(C)} Before applying capstan wraps, a mass tethered to MiniRHex pulls it off of the platform. MiniRHex demonstrates, with the use of a tether-wrapped capstan object, the ability to control the lowering and arresting of this same mass. \textbf{(D)} Two MiniRHex robots demonstrate their ability to apply tethers on rectangular posts to allow varied planar motion of a weighted dolly.}
    \label{fig:demos}
    \vspace{-6mm}
\end{figure}

\subsection{Laboratory demonstration}

Fig. 6 presents a summary of laboratory robotic demonstrations mirroring the modalities presented in Fig. 1. 

In Fig. 6A1, we demonstrate tethered payload locomotion in a laboratory environment containing a bed of granular media and two rock capstan objects.  
MiniRHex, in this low-traction environment, applies partial wraps of tether around two rocks, then acts as an anchor for the weighted dolly, 8x heavier than MiniRHex, to winch itself forward. In Fig. 6A2, the motion capture data from the granular media demonstration shows that, after the winch begins to pull on the tether, neither MiniRHex nor the rock capstans move. For the same two rock capstan objects, we utilize an unweighted sled and handheld force gauge to test the tension amplification capabilities of this environment (shown in Fig. 6B1). The resulting tension force data, plotted over time in Fig. 6B2, reaches an $A_F$ of 774 before one of the rock capstans dislodges from the substrate. No visually observable tether slip occurs during this trial.

In Fig. 6C, we affix a mass and tether to a MiniRHex robot in order to demonstrate controlled target transport load lowering. 
Before applying capstan wraps, the mass overcomes the static friction force between MiniRHex and the platform, and MiniRHex is pulled off the platform. After applying two wraps around a single capstan, MiniRHex then precisely controls the lowering and arresting of the motion of the same mass by either walking backwards or stopping, respectively.

In Fig. 6D, two MiniRHex’s tethered to the same weighted dolly apply tether wraps on two separate rectangular posts, which allows controlled planar motion of the dolly through two separately controlled on-board winches. 
We demonstrate that the dolly is able to utilize its winches individually for locomotion towards a single agent’s capstan object, or simultaneously to locomote in a direction between the two objects. The right tether snags under the table leg utilized in the lab, while the left tether does not.
These principles also translate into unstructured outdoor environments.

\begin{figure}[t]
    \centering
    \includegraphics[width=1\linewidth]{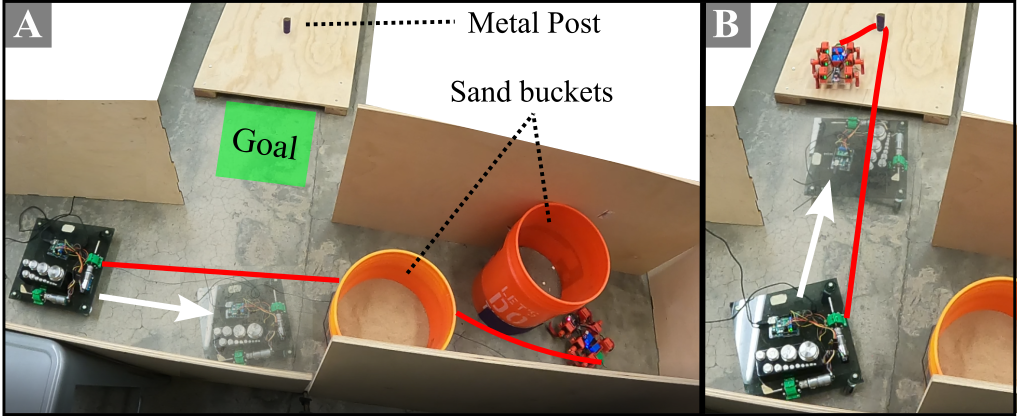}
    \setlength{\abovecaptionskip}{-10pt}
    \caption{ \textbf{Visual summary of sequential capstan maneuvers} MiniRHex applies partial wraps to two buckets to act as anchor for the dolly to winch towards. MiniRHex then unwraps the partial wraps and moves to wrap on a new capstan object, allowing the dolly to maneuver to the desired area.
    }
    \label{fig:Sequence}
    \vspace{-6mm}
\end{figure}

In Fig. \ref{fig:Sequence}, the dolly is now transported around an obstructive wall through multiple steps. This sequential task demonstrates the benefit of leveraging partial wraps on multiple capstan objects. 
 In the first step of the sequence, MiniRHex (Fig. 7A) applies low-angle partial wraps on two sand buckets with 180-grit sandpaper, used to mimic rock-tether contact. Multiple partial wraps are especially useful in this constrained space, where the MiniRHex is unable to fully encircle the buckets. The dolly then winches until located in a desirable position for the next maneuver. In Fig. 7B, MiniRHex then unwraps from the two buckets and applies a new wrap to the metal post. Once again, partial wraps provide an advantage, as opposed to full wraps or knots, because the tether path is reversed and re-applied elsewhere, without tangling. The dolly then winches a second time, moving to the desired location.

\subsection{Field demonstration}

In Fig. 8, a wheeled rover, stuck on a branch and unable to locomote, deploys a MiniRHex robot to apply a shared tether around an adjacent redwood tree. 
The MiniRHex successfully applies an approximately 360-degree wrap angle on the tree. The terrain consists of detritus, loose dirt and pine needles, yet the capstan-amplified holding force provides an anchor for the rover’s winch. The winch successfully frees the rover, which is 11x heavier than MiniRHex, from its stuck position. In this demonstration, slip occurs between the tree and tether, and the MiniRHex robot slips and tips over as the rover winches toward it. In a demonstration of non-catastrophic slipping, the maneuver succeeds, allowing the rover to 
free itself.
\begin{figure}[t]
    \centering
    \includegraphics[width=0.8\linewidth]{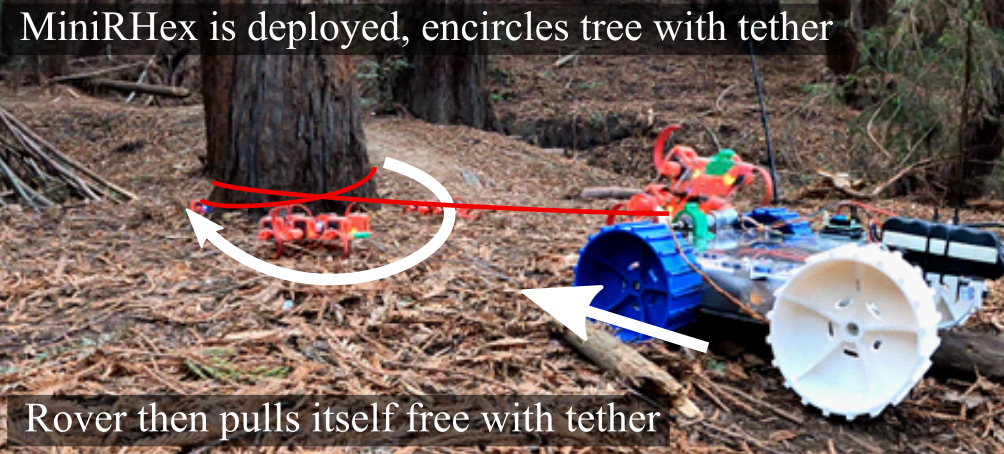}
    \setlength{\abovecaptionskip}{0pt}
    \caption{ \textbf{Visual summary of outdoor robotic demonstration}  MiniRHex in an outdoor environment applies its tether to act as an anchor point for a stuck rover, weighing over ten times its mass, to winch towards.}
    \label{fig:Roverstuck}
    \vspace{-6mm}
\end{figure}

\section{Discussion}
\label{sec:discuss}

Despite the variation of natural objects and tether configurations, we find that the idealized capstan equation matches experimental data from the field. Tether slip failure in capstan amplification is also not necessarily catastrophic, providing opportunities for successful maneuvering despite operating near the amplification factor limit. In addition, we show that harnessing the capstan effect across multiple objects provides advantages, including for maneuver reversibility. 



 


\subsection{Planning and executing tether deployment}


Tether deployment 
around natural capstans assumes the presence of suitable objects for wrapping, and 
Fig. 4 yields insight into the variety of usable objects on the UC Berkeley campus and a redwood forest. Less obvious surfaces for capstan amplification include bushes, curved cliff walls, and even buildings, assuming an ability to deploy such a long tether. The minimum acceptable size for a 
capstan object is a function of the radius of curvature of the tether. 


We envision generating a library of common capstan behaviors between different tethers and field objects across more expansive environments in future work. Characterizations, like that of redwood trees in Fig. \ref{fig:muinnature}B1, can guide path planning. The lowest measured friction coefficient and 95\% confidence interval represent factor of safety bounds for safety-critical and less dangerous maneuvers, respectively. These factors of safety then inform minimum required wrap angles for an operation. Further path planning and active sensing strategies
, in order to optimize the strength of multi-capstan systems, is a topic of future work. For example, risk of capstan object upheaval can be mitigated by advantageously dispersing the tether to multiple capstans to decrease the objects' individually-experienced external force.

The physical deployment of a tether, being dragged around an object, as in this work, is limited by the friction of the sliding tether and maximum pulling force of the tether-deploying agent. This problem can be mitigated if the agent uses a lightweight tether and overshoots, or runs past, the desired capstan object in order to begin its wrap with excess slack tether on the ground. Future work will investigate tether-deploying agents that can unspool tether locally during deployment to reduce tether-environment sliding, 
conceptually similar to tip-growing vines \cite{hawkes_soft_2017}, in order to engage with more objects or objects that are further apart.

\subsection{Entanglement and reversibility}

Entanglement can occur during the deployment of tethers around capstan objects, and affects both the strength of the maneuver and the ability to reverse tether deployment. For example, when entanglement occurs, functions like lowering the payload in Fig. \ref{fig:demos}C can no longer be achieved. Tethers, when deployed in a slack state first, have a propensity to self-entangle when cinched, especially at high wrap angles. Whenever the tether-deploying agent must cross over or under the portion of tether, 
it can also become entangled, such as hooking on a leg. When robot-tether entanglement occurs, the system enters a lasso- or knot-like envelopment grasp. 
After a maneuver is completed, entangled tether may need to be cut free and left behind, so tangles should be avoided if reversibility and reusability are desired. 
To avoid this circumstance, a distribution of partial wraps among multiple objects eases unwinding, especially if the tether never crosses its own path, as in Fig. 3A-d, \ref{fig:demos}A1 and \ref{fig:Sequence}. While removable knots present an option for reversible strong attachment, 
tether friction remains preferable for simplicity and necessary either for objects that cannot be encircled or for controlled sliding as in Fig. \ref{fig:demos}C.

\section{Conclusion}
\label{sec:conclusion}

Mobile robots can use natural capstans in unstructured environments as a way to bolster existing tractive force and achieve manipulations, with the addition of a tether attached to a payload or payload-winching agent. 
We specifically demonstrate features of this technique on a small-scale mobile system not optimized for forceful manipulation 
in order to highlight
the high load handling capability provided
with the simple addition of a tether and winch. This work therefore motivates the design and control of more specialized tethered systems that harness the capstan effect for specific applications. 

\subsection{Future work: Applications}

Equipped with a winch, urban search and rescue robots and remote explorers can use on-board deployable lightweight robots to traverse terrains that were previously unreachable, or to save a rover that has become stuck. Specialized lightweight robots could explore ahead, climbing steep inclinations, traversing loose substrates, or across unpredictable terrain, then act as anchor for a more massive robot or payload to winch itself across the obstruction. These same deployable robots could apply tethers near an excavation site, helping to move or lower massive raw material.
We expect that a variety of systems that can deploy a tether, whether flying, grounded, or submerged, can cooperate to achieve forceful maneuvers by harnessing this technique. 

\section*{Acknowledgement}

This work was supported by an Early Career Faculty grant from
NASA’s Space Technology Research Grants Program
(PI H.S.S., $\#$80NSSC21K0069). L.K.T was supported by a National Defense Science and Engineering Graduate Fellowship through the Office of Naval Research. Any opinions, findings, materials selection, and conclusions or recommendations expressed in this material are those of the author(s) and do not necessarily reflect the views of NASA/ONR.

\bibliography{bibtex/bib/4.20.bib}

\begin{thebibliography}{10}
\providecommand{\url}[1]{#1}
\csname url@samestyle\endcsname
\providecommand{\newblock}{\relax}
\providecommand{\bibinfo}[2]{#2}
\providecommand{\BIBentrySTDinterwordspacing}{\spaceskip=0pt\relax}
\providecommand{\BIBentryALTinterwordstretchfactor}{4}
\providecommand{\BIBentryALTinterwordspacing}{\spaceskip=\fontdimen2\font plus
\BIBentryALTinterwordstretchfactor\fontdimen3\font minus
  \fontdimen4\font\relax}
\providecommand{\BIBforeignlanguage}[2]{{%
\expandafter\ifx\csname l@#1\endcsname\relax
\typeout{** WARNING: IEEEtran.bst: No hyphenation pattern has been}%
\typeout{** loaded for the language `#1'. Using the pattern for}%
\typeout{** the default language instead.}%
\else
\language=\csname l@#1\endcsname
\fi
#2}}
\providecommand{\BIBdecl}{\relax}
\BIBdecl

\bibitem{estrada_forceful_2018}
M.~A. Estrada, S.~Mintchev \emph{et~al.}, ``Forceful manipulation with micro
  air vehicles,'' \emph{Science Robotics}, vol.~3, no.~23, p. eaau6903, 2018.

\bibitem{christensen_lets_2016}
D.~L. Christensen, S.~A. Suresh \emph{et~al.}, ``Let’s all pull together:
  Principles for sharing large loads in microrobot teams,'' \emph{{IEEE}
  Robotics and Automation Letters}, vol.~1, no.~2, pp. 1089--1096, 2016.

\bibitem{mcgarey_system_2016}
P.~{McGarey}, F.~Pomerleau, and T.~D. Barfoot, ``System design of a tethered
  robotic explorer ({TReX}) for 3d mapping of steep terrain and harsh
  environments,'' in \emph{Field and Service Robotics}.\hskip 1em plus 0.5em
  minus 0.4em\relax Springer, 2016, pp. 267--281.

\bibitem{stefanini_tether_2018}
A.~Stefanini and M.~Indri, ``Tether {Management} and {Tension} {Control} for
  {Rappelling} {Rovers},'' Master's thesis, Politecnico di Torino, 2018.

\bibitem{mcpherson_team-based_2019}
D.~{McPherson} and R.~S. Fearing, ``Team-based robot righting via pushing and
  shell design,'' in \emph{{IEEE} International Conference on Robotics and
  Automation ({ICRA})}, 2019, pp. 2168--2173.

\bibitem{deshpande_methodology_2009}
A.~D. Deshpande and J.~E. Luntz, ``A methodology for design and analysis of
  cooperative behaviors with mobile robots,'' \emph{Autonomous Robots},
  vol.~27, no.~3, pp. 261--276, 2009.

\bibitem{fukushima_development_2001}
E.~F. Fukushima, N.~Kitamura, and S.~Hirose, ``Development of tethered
  autonomous mobile robot systems for field works,'' \emph{Advanced Robotics},
  vol.~15, no.~4, pp. 481--496, 2001.

\bibitem{asano_tethered_2010}
N.~Asano, H.~Nakamoto \emph{et~al.}, ``Tethered detachable hook for the
  spiderman locomotion (design of the hook and its launching winch),'' in
  \emph{Field and Service Robotics}.\hskip 1em plus 0.5em minus 0.4em\relax
  Springer, 2010, pp. 25--34.

\bibitem{kitai_proposal_2005}
S.~Kitai, K.~Tsuru, and S.~Hirose, ``The proposal of swarm type wall climbing
  robot system “anchor climber”: the design and examination of adhering
  mobile unit,'' in \emph{International Conference on Intelligent Robots and
  Systems, {IEEE}/{RSJ}}, 2005, pp. 475--480.

\bibitem{shoval_design_1999}
S.~Shoval, E.~Rimon, and A.~Shapiro, ``Design of a spider-like robot for motion
  with quasi-static force constraints,'' in \emph{{IEEE} International
  Conference on Robotics and Automation}, vol.~2, 1999, pp. 1377--1383.

\bibitem{hao_design_2011}
G.~Hao, F.~Meng \emph{et~al.}, ``The design of cable-climbing robot,'' in
  \emph{2011 International Conference on Electronic and Mechanical Engineering
  and Information Technology ({EMEIT})}, vol.~4, 2011, pp. 2208--2211.

\bibitem{casarez_step_2016}
C.~S. Casarez and R.~S. Fearing, ``Step climbing cooperation primitives for
  legged robots with a reversible connection,'' in \emph{{IEEE} International
  Conference on Robotics and Automation ({ICRA})}, 2016, pp. 3791--3798.

\bibitem{mcgarey_developing_2018}
P.~{McGarey}, D.~Yoon \emph{et~al.}, ``Developing and deploying a tethered
  robot to map extremely steep terrain,'' \emph{Journal of Field Robotics},
  vol.~35, no.~8, pp. 1327--1341, 2018.

\bibitem{bares_dante_1999}
J.~Bares and D.~Wettergreen, ``Dante {II}: Technical description, results, and
  lessons learned,'' \emph{The International Journal of Robotics Research},
  vol.~18, no.~7, pp. 621--649, 1999.

\bibitem{mumm_planetary_2004}
E.~Mumm, S.~Farritor \emph{et~al.}, ``Planetary cliff descent using cooperative
  robots,'' \emph{Autonomous Robots}, vol.~16, no.~3, pp. 259--272, 2004.

\bibitem{schempf_self-rappelling_2009}
H.~Schempf, ``Self-rappelling robot system for inspection and reconnaissance in
  search and rescue applications,'' \emph{Advanced Robotics}, vol.~23, no.~9,
  pp. 1025--1056, 2009.

\bibitem{fernandez_tail-based_2021}
J.~Fernandez and A.~Mazumdar, ``Tail-based anchoring on granular media for
  transporting heavy payloads,'' \emph{{IEEE} Robotics and Automation Letters},
  vol.~6, no.~2, pp. 1232--1239, 2021.

\bibitem{schneider2021reachbot}
S.~Schneider, A.~Bylard \emph{et~al.}, ``Reachbot: A small robot for large
  mobile manipulation tasks,'' \emph{arXiv preprint arXiv:2110.10829}, 2021.

\bibitem{mcgarey_tslam_2017}
P.~Mcgarey, K.~{MacTavish} \emph{et~al.}, ``{TSLAM}: Tethered simultaneous
  localization and mapping for mobile robots,'' \emph{The International Journal
  of Robotics Research}, vol.~36, pp. 1363--1386, 2017.

\bibitem{kumar_entanglement_2007}
T.~Kumar and R.~Richardson, ``Entanglement detection of a swarm of tethered
  robots in search and rescue applications,'' in \emph{{ICINCO} 4th
  International Conference on Informatics in Control, Automation and Robotics,
  Proceedings}, vol.~2, 2007, pp. 143--148.

\bibitem{a_specian_friction_2015}
A.~Specian and M.~Yim, ``Friction binding study and remedy design for tethered
  search and rescue robots,'' in \emph{{IEEE} International Symposium on
  Safety, Security, and Rescue Robotics ({SSRR})}, 2015, p.~6.

\bibitem{augugliaro_building_2013}
F.~Augugliaro, A.~Mirjan \emph{et~al.}, ``Building tensile structures with
  flying machines,'' in \emph{IEEE/RSJ International Conference on Intelligent
  Robots and Systems}, 2013, pp. 3487--3492.

\bibitem{segal_method_2021}
E.~M. Segal, L.~Elkowitz, and N.~Belitsis, ``A method for rapidly deploying
  suspended footbridges,'' \emph{Structural Engineering International},
  vol.~31, no.~3, 2021.

\bibitem{d_kuhm_fabric_2014}
D.~Kuhm, M.-A. Bueno, and D.~Knittel, ``Fabric friction behavior: study using
  capstan equation and introduction into a fabric transport simulator,''
  \emph{Textile Research Journal}, vol.~84, pp. 1070--1083, 2014.

\bibitem{howell_friction_1954}
H.~G. Howell, ``The friction of a fibre round a cylinder and its dependence
  upon cylinder radius,'' \emph{Journal of the Textile Institute Transactions},
  vol.~45, no.~8, 1954.

\bibitem{jayawardana_capstan_2021}
K.~Jayawardana, ``\BIBforeignlanguage{en}{Capstan {Equation} {Generalised} for
  {Noncircular} {Geometries}},'' \emph{\BIBforeignlanguage{en}{arXiv preprint
  arXiv.2012.13353}}, Dec. 2020.

\bibitem{jiang_stochastic_2018}
H.~Jiang, S.~Wang, and M.~R. Cutkosky, ``Stochastic models of compliant spine
  arrays for rough surface grasping,'' \emph{The International Journal of
  Robotics Research}, vol.~37, no.~7, pp. 669--687, 2018.

\bibitem{yano_development_1997}
T.~Yano, T.~Suwa \emph{et~al.}, ``Development of a semi self-contained wall
  climbing robot with scanning type suction cups,'' in \emph{{IEEE}/{RSJ}
  International Conference on Intelligent Robots and Systems}, vol.~2, 1997,
  pp. 900--905.

\bibitem{prahlad_electroadhesive_2008}
H.~Prahlad, R.~Pelrine \emph{et~al.}, ``Electroadhesive robots—wall climbing
  robots enabled by a novel, robust, and electrically controllable adhesion
  technology,'' in \emph{{IEEE} International Conference on Robotics and
  Automation}, 2008, pp. 3028--3033.

\bibitem{wang_large-payload_2013}
L.~Wang, L.~Graber, and F.~Iida, ``Large-payload climbing in complex vertical
  environments using thermoplastic adhesive bonds,'' \emph{{IEEE} Transactions
  on Robotics}, vol.~29, no.~4, pp. 863--874, 2013.

\bibitem{hawkes_geckos_2013}
E.~W. Hawkes, E.~V. Eason \emph{et~al.}, ``The gecko’s toe: Scaling
  directional adhesives for climbing applications,'' \emph{{IEEE}/{ASME}
  Transactions on Mechatronics}, vol.~18, no.~2, pp. 518--526, 2013.

\bibitem{roderick_bird-inspired_nodate}
W.~R.~T. Roderick, M.~R. Cutkosky, and D.~Lentink, ``Bird-inspired dynamic
  grasping and perching in arboreal environments,'' \emph{Science Robotics},
  vol.~6, no.~61, p. eabj7562, 2022.

\bibitem{thomas_visual_2016}
J.~Thomas, G.~Loianno \emph{et~al.}, ``Visual servoing of quadrotors for
  perching by hanging from cylindrical objects,'' \emph{{IEEE} Robotics and
  Automation Letters}, vol.~1, no.~1, pp. 57--64, 2016.

\bibitem{meder_plant_2022}
F.~Meder, S.~P. Murali~Babu, and B.~Mazzolai, ``A {Plant} {Tendril}-{Like}
  {Soft} {Robot} {That} {Grasps} and {Anchors} by {Exploiting} its {Material}
  {Arrangement},'' \emph{IEEE Robotics and Automation Letters}, vol.~7, no.~2,
  pp. 5191--5197, Apr. 2022.

\bibitem{green_wood_2001}
D.~W. Green, ``Wood: Strength and stiffness,'' in \emph{Encyclopedia of
  Materials: Science and Technology}.\hskip 1em plus 0.5em minus 0.4em\relax
  Elsevier Science Ltd., 2001, pp. 9732--9736.

\bibitem{wyllie_chapter_1996}
D.~C. Wyllie and N.~I. Norrish, ``Chapter 14 - rock strength properties and
  their measurement,'' in \emph{Landslides: Investigation and Mitigation},
  1996, pp. 372-- 389.

\bibitem{lindroos_forces_2010}
O.~Lindroos, M.~Henningsson \emph{et~al.}, ``Forces required to vertically
  uproot tree stumps,'' \emph{Silva Fennica}, vol.~44, 2010.

\bibitem{li_terradynamics_2013}
C.~Li, T.~Zhang, and D.~I. Goldman, ``A terradynamics of legged locomotion on
  granular media,'' \emph{Science}, vol. 339, no. 6126, pp. 1408--1412, 2013.

\bibitem{barragan_minirhex_2018}
M.~Barragan, N.~Flowers, and A.~M. Johnson, ``{MiniRHex}: A small, open-source,
  fully programmable walking hexapod,'' \emph{Robotics: Science and Systems
  Workshop on “Design and Control of Small Legged Robots”}, p.~2, 2018.

\bibitem{cao_mobility_2021}
C.~Cao, C.~Creager \emph{et~al.}, ``Mobility experiments assessing performance
  of front-back differential drive velocity on sandy terrain,'' in \emph{20th
  International and 9th Americas Conference of the {ISTVS}}, 2021.

\bibitem{hawkes_soft_2017}
E.~W. Hawkes, L.~H. Blumenschein \emph{et~al.}, ``A soft robot that navigates
  its environment through growth,'' \emph{Science Robotics}, vol.~2, no.~8, p.
  eaan3028, 2017.

\end{thebibliography}
\bibliographystyle{IEEEtran}

\end{document}